\documentclass{article}
\usepackage{spconf,amsmath,graphicx}
\usepackage{amssymb}
\usepackage[nohyperlinks]{acronym}
\usepackage[disable,colorinlistoftodos]{todonotes}
\usepackage{bm}
\usepackage[table]{xcolor}
\usepackage{booktabs}
\usepackage{multirow}
\usepackage{algorithm}
\usepackage{algorithmic}
\usepackage{tikz}
\usetikzlibrary{arrows.meta,positioning}
\definecolor{LinkBlue}{RGB}{0,70,160}
\definecolor{CiteBlue}{RGB}{0,90,170}
\definecolor{UrlBlue}{RGB}{0,90,200}

\usepackage[colorlinks=true, linkcolor=LinkBlue, citecolor=CiteBlue, urlcolor=UrlBlue]{hyperref}
\usepackage{cleveref}
\usepackage{soul}
\setuldepth{foobar}
\usepackage{graphicx}
\usepackage{subcaption}

\crefname{algorithm}{Alg.}{Algs.}
\crefname{equation}{Eq.}{Eqs.}
\crefname{figure}{Fig.}{Figs.}
\crefname{table}{Table}{Tables}
\crefname{section}{Section}{Sections}
\crefname{subsection}{Section}{Sections}

\def\modelName{EASE}
\def\modelNameLong{\textbf{E}ncoder-only \textbf{A}udio-Visual \textbf{Se}gmentation}
\acrodef{\modelName}{\modelNameLong}
\acrodef{AVS}{Audio-Visual Segmentation}
\acrodef{AVSS}{Audio-Visual Semantic Segmentation}
\acrodef{SotA}{State-of-the-Art}
\acrodef{S4}{Single Sound Source Segmentation}
\acrodef{MS3}{Multiple Sound Source Segmentation}
\acrodef{ViT}{Vision Transformer}
\acrodef{VFM}{Vision Foundation Models}
\acrodef{PVTv2}{Pyramid Vision Transformer v2}
\acrodef{SSL}{Sound Source Localization}
\acrodef{CNN}{Convolutional Neural Network}

\usepackage{xspace}

\def\eqref#1{equation~\ref{#1}}

\def\1{\bm{1}}

\def\vc{{\bm{c}}}

\def\vf{{\bm{f}}}

\def\vq{{\bm{q}}}

\def\vs{{\bm{s}}}

\DeclareMathAlphabet{\mathsfit}{\encodingdefault}{\sfdefault}{m}{sl}
\SetMathAlphabet{\mathsfit}{bold}{\encodingdefault}{\sfdefault}{bx}{n}

\def\sR{{\mathbb{R}}}

\newcommand{\R}{\mathbb{R}}

\newcommand{\PAR}[1]{\vskip4pt
\noindent
{\bf #1~}}
\newcommand{\PARbegin}[1]{\noindent
{\bf #1~}}

\renewcommand{\paragraph}[1]{\vspace{.5em}
\noindent
\textbf{#1.}}

\usepackage{microtype}

\newcommand{\tikzxmark}{%
\tikz[scale=0.23]{ \draw[line width=0.7,line cap=round] (0,0) to [bend left=6] (1,1);
\draw[line width=0.7,line cap=round] (0.2,0.95) to [bend right=3] (0.8,0.05); }}
\newcommand{\tikzcmark}{%
\tikz[scale=0.23]{ \draw[line width=0.7,line cap=round] (0.25,0) to [bend left=10]
(1,1); \draw[line width=0.8,line cap=round] (0,0.35) to [bend right=1] (0.23,0);
}}

\title{Less is More: \uppercase\expandafter{\modelNameLong}}

\name{Ilpo Viertola$^{1}$\thanks{We acknowledge the Finnish IT Center for Science (CSC) for computational resources.}, Vladimir Iashin$^{1}$, Sophie Tötterström$^{1}$, Esa Rahtu$^{1}$}
\address{$^{1}$Tampere University}

\begin{document}
\ninept
\maketitle
\begin{abstract}
\acf{AVSS} aims to identify, segment, and classify sound-emitting objects in video frames.  
Previous Transformer-based \ac{AVSS} approaches largely inherit design principles from image segmentation models.
Recent studies show that these image segmentation models contain redundant components that contribute little to the segmentation performance.
Following this insight, we propose \ac{\modelName}.
\modelName~runs at up to 365 FPS, 3$\times$ faster than prior \ac{SotA} AVS models at comparable accuracy, and trains in under 11 GPU-hours.
Furthermore, we achieve \ac{SotA} \ac{AVSS} performance across different backbones and input resolutions. 
Our results demonstrate that \ac{AVSS} can be both simpler and faster, providing a scalable foundation for future research and real-time applications. 
Code, model weights, and samples are available at \href{https://ease-avs.notion.site/}{ease-avs.notion.site}.
\end{abstract}
\begin{keywords}
Semantic Segmentation, Audio-Visual Learning
\end{keywords}
\section{Introduction}
\label{sec:intro}

\acf{AVS} aims to localize and segment sounding objects in a video.
Traditionally, the \ac{AVS} task is divided into three sub-tasks: \acf{S4}, \acf{MS3}, and \acf{AVSS} \cite{zhou2022audio, zhou2024avss}.
For the first two, the goal is to produce a pixel-level binary mask indicating pixels emitting the sounds.
For the semantic segmentation, the goal is also to categorize the masked objects.
Since \ac{AVSS} is larger in size, a more complex task, and includes all samples from \ac{S4} and \ac{MS3}, we focus on it in this work.

\begin{figure}[ht]
    \centering
    \begin{subfigure}
        [b]{0.45\textwidth}
        \centering
        \caption{Common Transformer-based AVS approach.}
        \vspace{2pt}
        \includegraphics[width=\textwidth]{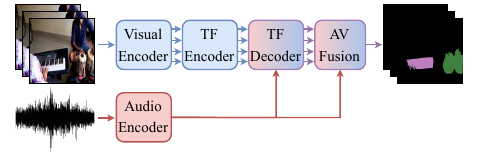}
        \hspace{-13pt} \label{fig:common_appr}
    \end{subfigure}
    \hfill
    \vspace{-3pt}
    \begin{subfigure}
        [b]{0.45\textwidth}
        \centering
        \caption{Our \acf{\modelName}}
        \vspace{2pt}
        \includegraphics[width=\textwidth]{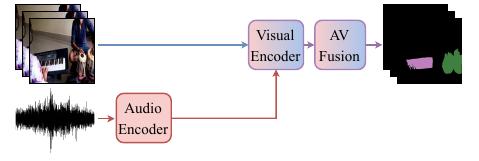}
        \label{fig:aves_simple}
    \end{subfigure}
    \vspace{-10pt}
    \caption{
        \textbf{Comparison between our simplified approach and common Transformer-based AVS methods.}
        Our approach eliminates unnecessary architectural complexity, resulting in a more efficient model while preserving high segmentation performance. 
    }
    \label{fig:combined_arc}
\end{figure}

Many AVS studies draw inspiration from image segmentation research \cite{gao2024avsegformer, wang2024avesformer, yang2024cooperation, shi2024cross, ma2024stepping, li2024qdformer, li2024selm}. 
Recent works \cite{kerssies2025your, chen2021simple, fang2021you} have explored redundant components in current \acf{SotA} image segmentation models, while the plain \ac{ViT} \cite{dosovitskiy2020vit} has demonstrated strong generalization across a wide range of computer vision tasks \cite{oquab2023dinov2, simeoni2025dinov3, dehghani2023scaling}.
Additional task-specific components implemented on top of a \ac{ViT} backbone in image segmentation models often increase model size and computational cost without substantially improving segmentation quality, motivating efforts towards more efficient and modular architectures \cite{kerssies2025your, chen2021simple, fang2021you}. 

A common Transformer-based \cite{vaswani2017attention} AVS approach is illustrated in \cref{fig:common_appr}. 
We adapt the core design principles of \cite{kerssies2025your, chen2021simple, fang2021you} to \ac{AVS} by removing task-specific building blocks in favor of a plain \ac{ViT} architecture, and introduce \acf{\modelName}.
Our approach is shown in \cref{fig:aves_simple}.
It enables integrating \ac{VFM}s \cite{simeoni2025dinov3, oquab2023dinov2} to leverage their strong visual representations while reducing complexity.
We experiment with two visual backbones: the commonly used \ac{AVS} backbone \ac{PVTv2} \cite{wang2022pvt}, and the widely adopted \ac{VFM} DINOv3 \cite{simeoni2025dinov3}. 
Thanks to extensive pretraining and high \ac{VFM} capacity, we achieve \ac{SotA} \ac{AVSS} performance with $3\times$ higher throughput. 
The scalable vanilla \ac{ViT} \cite{dosovitskiy2020vit} architecture further enables optimized attention implementations such as FlashAttention \cite{dao2023flashattention2} for faster training and inference.

Our contributions can be summarized as follows: i) a simple, strong, and efficient baseline for \ac{AVS} that enables adopting \ac{VFM}s for \ac{AVS}, ii) a model-agnostic cross-modal fusion technique with audio feature enhancement, and iii) extensive experiments on all \ac{AVS} sub-tasks to demonstrate the balance between segmentation performance and efficiency of our method.


\section{Background}
\label{sec:background}
Audio-visual segmentation (AVS) refers to the task of predicting pixel-wise masks for the sound-producing objects in a video.
It was first proposed by Zhou et al. \cite{zhou2022audio, zhou2024avss}, and it is closely related to \ac{SSL} \cite{arandjelovic2017look, arandjelovic2018objects, senocak2018learning}.
\ac{AVS} consists of three sub-tasks, each with a corresponding dataset: \acf{S4}, \acf{MS3}, and \acf{AVSS}. 
Since the first two concentrate on binary predictions (sound-producing vs. silent pixel), we mainly focus on the \ac{AVSS} task, where each pixel is also assigned a class label. 
\ac{S4} and \ac{MS3} data are subsets of \ac{AVSS}.

Together with the datasets, Zhou et al. \cite{zhou2022audio, zhou2024avss} introduced a \acf{CNN} based temporal pixel-wise audio-visual interaction (TPAVI) network. 
Later AVS methods utilize cross-attention \cite{vaswani2017attention} and use audio as a reference query \cite{gao2024avsegformer, wang2024avesformer, shi2024cross, ma2024stepping, huang2023discovering, chen2024unraveling, li2023catr, liu2024bavs} to fuse cross-modal information. 
To emphasize audio, recent works have introduced bidirectional attention, where audio is used to compute the key and value vectors \cite{yang2024cooperation, chen2024bootstrapping}.

Many of the aforementioned approaches draw inspiration from Mask Transformer \cite{carion2020end}. 
Adaptation of the Mask Transformer-based approach for \ac{AVS} relies on semantic alignment of visual and auditory features \cite{yang2024cooperation, ma2024stepping, li2023catr, chen2024bootstrapping, liu2023audio}. Alternatively, Gong et al. \cite{gong2025avs} explore selective state-space models for \ac{AVS}. 
Mao et al. \cite{mao2023multimodal, mao2025contrastive} explore modeling a shared modality space using a conditional variational autoencoder \cite{kingma2013auto} and a diffusion model \cite{ho2020denoising} to learn an effective multimodal latent space for segmentation.

Many \ac{AVS} models are built on top of image segmentation models.
As a result, they inherit the same architectural components: hierarchical backbones, multi-scale feature pyramids, and localized attention. 
These components add task-specific complexity to the plain \ac{ViT} architecture utilized by modern \ac{VFM}s.
Because \ac{AVS} models deviate from plain \ac{ViT} design, they cannot benefit from \ac{VFM} pretraining without significant re-engineering.
Kerssies et al. \cite{kerssies2025your} demonstrate this incompatibility in the image segmentation domain. 
This motivates us to build a simpler, efficient, and scalable alternative to specialized architectures in the audio-visual segmentation domain.


\section{Method}
\label{sec:method}
\begin{figure}[t]
    \centering
    \includegraphics[]{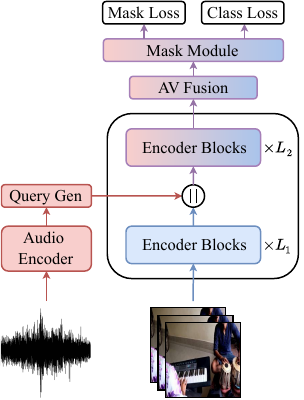}
    \caption{ 
        \textbf{Overview of \modelName.}
        Audio and visual frames are encoded, and the encoded audio is enhanced by soft clustering it to semantically representative auditory centers. 
        The query generator is used to generate sparse audio queries before concatenating them with the visual features.
        The concatenated sequence is processed with the remaining $L_{2}$ encoder blocks and fused at each block with the enhanced audio features. 
        The connection between enhanced audio features and AV Fusion is omitted for clarity. 
        Finally, a mask module is employed to produce the final mask predictions from the visual tokens and the class predictions from the sparse audio query tokens. 
    }
    \label{fig:aves}
\end{figure}

\acf{\modelName}, presented in \cref{fig:aves}, is a simplified approach for \ac{AVS} built on top of a plain \ac{ViT} architecture making it fully compatible with modern \ac{VFM}s.
It also introduces new components: a Gumbel-Softmax-based audio feature enhancement module, a mid-backbone audio query injection strategy, and a bidirectional audio-visual fusion module with learned soft clustering.
Our model achieves state-of-the-art performance while being up to $3\times$ faster than existing methods.

\PAR{Audio Feature Extraction.}
We follow the standard pipeline established in~\cite{zhou2022audio,zhou2024avss}.
Audio is encoded with a pretrained and frozen VGGish~\cite{hershey2017cnn} encoder, producing frame-level features $\vf_{a}\in \sR^{T \times 128}$, where $T$ is the number of frames. 
This yields a single feature vector per frame, which lacks discriminative semantic structure across different sound sources and intra-class variations.

To address this, we introduce a clustering step that groups audio features around $K$ learned semantic centers $\vc_{a}\in \sR^{T \times K \times 128}$. 
Each frame is assigned to the centers based on feature similarity, producing audio representations that better capture semantic variation within and across sound categories. 
We refer to this process as \textit{audio feature enhancement}.
In our experiments, setting $K$ to the number of semantic classes in AVSBench~\cite{zhou2022audio, zhou2024avss} yielded the best results.

The cluster assignments are learned end-to-end via the Gumbel-Softmax operation, which enables discrete-like cluster selection while remaining fully differentiable during training. 
Standard softmax produces soft, distributed assignments across all $K$ centers, whereas Gumbel-Softmax encourages sparser, more committed assignments that correspond to distinct semantic categories.
Following \cite{liu2025dynamic}, we use $\vc_{a}$ to compute a discriminative semantic enhancement for $\vf_{a}$, which reduces semantic bias arising from intra-class variation. 
The resulting enhanced audio feature is denoted $\hat{\vf}_{a} \in \sR^{T \times K \times 128}$.

Our approach differs from DDESeg~\cite{liu2025dynamic}, which precomputes the audio feature bank prior to training.
By learning the cluster centers and assignments end-to-end, our method eliminates dataset-level precomputation and adaptively captures the semantic structure of the audio, improving the discriminative power of the features for \ac{AVS}.

\PAR{Mask Prediction.}
The mask prediction design adopts the plain \ac{ViT} segmentation approach of Kerssies et al. \cite{kerssies2025your} as its structural foundation. 
Our contribution is the mid-backbone audio query injection strategy, which adapts their design to the multimodal \ac{AVS} setting.

\ac{\modelName} employs a \ac{ViT}-based \cite{dosovitskiy2020vit} visual backbone to encode input video frames.
Rather than processing visual and audio tokens jointly from the start, we split the backbone into two stages at block $L_{1}$. 
The first stage processes visual tokens independently, producing rich visual representations $\vf_{m} \in \sR^{T \times N_m \times D}$, where $N_{m}$ is the number of patch tokens, and $D$ is the shared feature dimension.

After $L_{1}$ blocks, we generate sparse audio queries from the enhanced audio feature $\hat{\vf}_{a}$ using the query generator (\textit{Query Gen} in \cref{fig:aves}).
This is done by cross-attending $\hat{\vf}_{a}$ with a learnable query vector, using $\hat{\vf}_{a}$ as keys and values.
The audio representation is then projected to a shared feature dimension $D$ of the visual backbone, yielding a set of sparse queries $\vq \in \sR^{T \times N_q \times D}$, where $N_{q}$ is the number of query tokens. 
Since $\vf_{m}$ already carries positional information from the $L_{1}$ stage, learnable positional encodings are added only to $\vq$. 
The queries are then concatenated with the visual token sequence along the sequence dimension to form the joint cross-modal sequence $\vs^{(0)}= \left[\, \vq \;;\; \vf_{m}\,\right] \in \sR^{T \times (N_q + N_m) \times D}$.

At each block $l \in \{L_{1}+1, \ldots, L_{2}\}$, we first split $\vs^{(l-1)}$ back into its constituent sequences $\vq^{(l-1)}$ and $\vf_m^{(l-1)}$, apply the audio-visual fusion module to the visual tokens using $\hat{\vf}_{a}$, and re-form the joint sequence before passing it through the ViT block.
All $\mathrm{AVFusion}$ modules across $L_{2}$ blocks share weights, keeping the added parameter count negligible.

Mask predictions are produced directly from $\vs^{(L_2)}$. 
Following \cite{kerssies2025your,cheng2022masked}, a small mask module predicts the final segmentation masks from $\vf_{m}^{L_2}$ and the class labels from $\vq^{L_2}$.
During training, we also apply this module after each $\{L_{1}+1, \ldots, L_{2}-1\}$ block to produce intermediate masks that constrain the self-attention of $\vq^{(l)}$, providing localization supervision.
Since masked attention is expensive, we follow \cite{kerssies2025your} and anneal it during training.

\PAR{Audio-Visual Fusion Module.}
The $\mathrm{AV~Fusion}$ module takes visual mask features $\vf_{m}^{(l-1)}$ and enhanced audio features $\hat{\vf}_{a}$ as input and returns enhanced visual tokens $\tilde{\vf}_{m}^{(l)}$. 
It operates only on token sequences of dimension $D$, so it imposes no constraints on the visual backbone and applies to any \ac{ViT}-based model without modification. 
Prior to fusion, $\hat{\vf}_{a}$ is projected to the shared dimension $D$.

\PAR{\textit{Stage 1: Visual-Guided Audio Suppression}} 
We cluster the visual tokens using Gumbel-Softmax clustering, with $K_v$ learnable visual cluster centers, producing visual clusters $\vc_{m} \in \R^{T \times K_v \times D}$. 
In our experiments, $K_v=5$ yielded the best results.
We then attend $\hat{\vf}_{a}$ over $\vc_{m}$ via cross-attention to assess how well each audio cluster is grounded in the visual scene, and map the result to a score $\vs \in [0,1]^{T \times K_v}$ through a linear layer and sigmoid operation.
Scaling $\hat{\vf}_{a}$ with $\vs$ yields visually-grounded audio features $\tilde{\vf}_{a}$ mitigating the interference of off-screen sounds.

\PAR{\textit{Stage 2: Audio-Conditioned Visual Feature Update}}
We condition the visual tokens on $\tilde{\vf}_{a}$ via a second cross-attention, with $\vf_{m}^{(l-1)}$ as queries and $\tilde{\vf}_{a}$ as keys and values, producing the updated visual tokens $\tilde{\vf}_{m}^{(l)}$. 
These are re-concatenated with $\vq^{(l)}$ and processed through the next \ac{ViT} block.


\section{Experiments}
\label{sec:experiments}
\subsection{Setup}
\label{ssec:setup}

\PARbegin{Datasets.}
We utilize the widely adopted audio-visual segmentation dataset AVSBench \cite{zhou2022audio, zhou2024avss}. 
The complete dataset includes three subsets: \ac{S4}, \ac{MS3}, and \ac{AVSS}. 
Videos are 10 seconds long, with one frame per second extracted for segmentation.
Input data is uniformly reshaped to $224 \times 224$ ($224^2$) or $384 \times 384$ ($384^2$).  
Following prior works, masks are predicted in $224 \times 224$ spatial size.

\PAR{Implementation.}
We utilize pretrained \ac{PVTv2} \cite{wang2022pvt} and DINOv3 \cite{simeoni2025dinov3} as visual backbone networks. 
While \ac{PVTv2} is widely adopted in AVS research, our simplified model design enables the effective use of \ac{VFM}s like DINOv3 with a modest computation budget. 
For \ac{PVTv2}, we use \textit{Stage 3} visual tokens and disable spatial reduction for $L_{2}$ blocks to improve efficiency.

In our experiments, $L_{2}=6$ and $L_{2}=4$ yielded the best results for \ac{PVTv2} and DINOv3 respectively. 
For DINOv3 training, we follow \cite{kerssies2025your}, including the loss computation. 
For \ac{PVTv2}, we adopt the same training strategy but disable the layer-wise learning rate decay to enable full model fine-tuning.
For training and testing, we utilize two NVIDIA A100 GPUs. 
We train \ac{S4} for 46 and 22 epochs, \ac{MS3} for 80 and 40 epochs, and \ac{AVSS} for 36 and 22 epochs for \ac{PVTv2} and DINOv3 models, respectively.
In GPU hours, training takes up to 7 and 5 hours for \ac{S4}, up to 1.5 hours for both for \ac{MS3}, and up to 15 and 11 hours for \ac{AVSS} using the \ac{PVTv2} and DINOv3.

All methods, except DDESeg \cite{liu2025dynamic}, use the frozen VGGish \cite{hershey2017cnn} audio encoder (72M parameters) as their audio backbone. 
DDESeg \cite{liu2025dynamic} replaces VGGish \cite{hershey2017cnn} with HTSAT, a substantially smaller and more modern audio encoder (29M parameters). 
We report parameter counts, including the audio encoder, to ensure transparency in \cref{tab:eff_study}.

\PAR{Evaluation Metrics.} 
For quantitative evaluation, we follow common practice and use the mean Jaccard index ($\mathcal{J}$) \cite{everingham2010pascal} and F-score with $\beta^{2}=0.3$ ($\mathcal{F}$).
To evaluate the model efficiency, we utilize inference speed frames per second (FPS) and the average number of floating-point operations (FLOPs) calculated with a batch size of 1 on a single NVIDIA RTX 4090 GPU. 
FLOPs are obtained using the DeepSpeed \cite{rasley2020deepspeed} profiler and reported as GFLOPs (FLOPs $\times 10^{9}$).
Note that reported GLOPs are not operations per second but rather the number of floating-point operations. 
We evaluate all models using their official codebases while employing a unified dataloading pipeline to eliminate differences in dataloading efficiency.

\subsection{Main Results}
\label{ssec:main_res}
\begin{table}[t]
    \centering
    \renewcommand{\tabcolsep}{1.15pt}
    \caption{
    \textbf{Quantitative comparison on \ac{AVSS} \cite{zhou2022audio, zhou2024avss} with efficiency evaluations.}
    We compare performance on \ac{AVSS} and throughput between \ac{\modelName} and \ac{SotA} approaches with open-source code.
    $^\dagger$Retrained on the standard data split and benchmark metrics.
    $^\ddagger$DDESeg uses HTSAT (29M) as its audio encoder, while all other methods use VGGish (72M) \cite{hershey2017cnn}. Parameter counts are reported inclusive of audio encoders.
    }
    \vspace{-5pt}
    \begin{tabular}{llcc ccccc}
        \toprule
        \multicolumn{1}{c}{} &
        \multicolumn{1}{c}{} &
        \multicolumn{1}{c}{} &
        \multicolumn{1}{c}{} & &
        \multicolumn{4}{c}{\textit{AVSS}} \\
        \cmidrule{6-9}
        \multicolumn{1}{l}{\multirow{-2}{*}{Method}} &
        \multicolumn{1}{l}{\multirow{-2}{*}{Backb.}} &
        \multicolumn{1}{l}{\multirow{-2}{*}{Params}} &
        \multicolumn{1}{l}{\multirow{-2}{*}{Inp.}} & &
        $\mathcal{J} \uparrow$ & $\mathcal{F} \uparrow$ & FPS$\uparrow$ & GFLOPs$\downarrow$ \\
        \midrule
        AVSegFormer \cite{gao2024avsegformer} & PVTv2 & 186M & $224^2$ &&  36.7 & 42.0 & 27 & 99 \\
        AVSegFormer \cite{gao2024avsegformer} & PVTv2 & 186M & $512^2$ && 37.3 & 42.8 & 23 & 504 \\
        AAVS \cite{ma2024stepping} & Swin-B & 187M & $384^2$ && 48.5 & 53.2 & 69 & 151 \\
        Selm \cite{li2024selm} & Swin-B & 186M & $448^2$ && 41.3 & 46.9 & 71 & 258 \\
        COMBO \cite{yang2024cooperation} & PVTv2 & 499M & $224^2$ && 42.1 & 46.1 & 73 & 346 \\
        DDESeg$^\dagger$$^\ddagger$ \cite{liu2025dynamic} & Transf. & 133M & $224^2$ && 43.3 & 48.7 & 27 & 179 \\
        \midrule
        \textbf{\modelName} & PVTv2 & 147M & $224^2$ && 42.3 & 47.6 & 159 & \textbf{40} \\
        \textbf{\modelName} & PVTv2 & 147M & $384^2$ && 46.1 & 50.6 & 141 & 105 \\
        \textbf{\modelName} & ViT-B & 187M & $224^2$ && 45.8 & 50.7 & \textbf{365} & \underline{96} \\
        \textbf{\modelName} & ViT-B & 187M & $384^2$ && 49.8 & 54.1 & 181 & 291 \\
        \textbf{\modelName} & ViT-L & 424M & $224^2$ && \underline{52.2} & \underline{57.2} & \underline{225} & 227 \\
        \textbf{\modelName} & ViT-L & 424M & $384^2$ && \textbf{56.5} & \textbf{61.0} & 86 & 721 \\
        \bottomrule
    \end{tabular}
    \label{tab:eff_study}
\end{table}


\begin{table}[t]
\centering
\renewcommand{\tabcolsep}{1.5pt}
\caption{
    \textbf{Quantitative comparison on \ac{S4}, \ac{MS3} and \ac{AVSS} categorized by backbone model and input size \cite{zhou2022audio, zhou2024avss}.}
    *\ac{MS3} model was pretrained on \ac{S4} data.
    $^\dagger$Retrained using standard data split and benchmark metrics.
    $^\ddagger$DDESeg uses HTSAT as its audio encoder, while all other methods use VGGish \cite{hershey2017cnn}.
}
\vspace{-5pt}
\begin{tabular}{llc c cc c cc c cc}
\toprule
\multicolumn{1}{c}{} &
\multicolumn{1}{c}{} &
\multicolumn{1}{c}{} & &
\multicolumn{2}{c}{\textit{AVSS}} & &
\multicolumn{2}{c}{\textit{S4}} & &
\multicolumn{2}{c}{\textit{MS3}} \\
\cmidrule{5-6}
\cmidrule{8-9}
\cmidrule{11-12}

\multicolumn{1}{l}{\multirow{-2}{*}{Method}} &
\multicolumn{1}{l}{\multirow{-2}{*}{Backb.}} &
\multicolumn{1}{l}{\multirow{-2}{*}{Inp.}} & &
$\mathcal{J} \uparrow$ &
$\mathcal{F} \uparrow$ & &
$\mathcal{J} \uparrow$ &
$\mathcal{F} \uparrow$ & &
$\mathcal{J} \uparrow$ &
$\mathcal{F} \uparrow$ \\
\midrule

TPAVI \cite{zhou2022audio} & PVTv2 & $224^2$ & & 29.8 & 35.2 & & 78.7 & 87.9 & & 54.0 & 64.5 \\
CATR \cite{li2023catr} & PVTv2 & $224^2$ & & 32.8 & 38.5 & & \underline{84.4} & \underline{91.3} & & \textbf{62.7} & \textbf{74.5} \\
ECMVAE \cite{mao2023multimodal} & PVTv2 & $224^2$ & & - & - & & 81.7 & 90.1 & & 57.8 & 70.8 \\
AQFormer* \cite{huang2023discovering} & PVTv2 & $224^2$ & & - & - & & 81.6 & 89.4 & & \underline{62.2} & \underline{72.7} \\
AVSegFormer \cite{gao2024avsegformer} & PVTv2 & $224^2$ & & 36.7 & 42.0 & & 82.1 & 89.9 & & 58.4 & 69.3 \\
COMBO \cite{yang2024cooperation} & PVTv2 & $224^2$ & & \underline{42.1} & \underline{46.1} & & \textbf{84.7} & \textbf{91.9} & & 59.2 & 71.2 \\
CAVP \cite{chen2024unraveling} & PVTv2 & $224^2$ & & 30.4 & 35.3 & & 78.8 & 88.9 & & 55.8 & 67.1 \\
\textbf{\modelName}* & PVTv2 & $224^2$ & & \textbf{42.3} & \textbf{47.6} & & 80.4 & 89.7 & & 60.8 & 68.7 \\

\midrule

AVSegFormer \cite{gao2024avsegformer} & PVTv2 & $512^2$ & & 37.3 & 42.8 & & 83.1 & 90.5 & & 61.3 & \underline{73.0} \\
AVS-Mamba \cite{gong2025avs} & PVTv2 & $448^2$ & & 39.7 & 45.1 & & \textbf{85.0} & \textbf{92.6} & & \textbf{68.6} & \textbf{78.6} \\
Selm \cite{li2024selm} & PVTv2 & $448^2$ & & \underline{41.3} & \underline{46.9} & & \underline{83.5} & \underline{91.2} & & 60.3 & 71.3 \\
\textbf{\modelName}* & PVTv2 & $384^2$ & & \textbf{46.1} & \textbf{50.6} & & 83.0 & 90.9 & & \underline{61.9} & 69.8 \\

\midrule

AVSC \cite{liu2023audio} & Swin-B & $224^2$ & & - & - & & 81.3 & 88.6 & & 59.5 & 65.7 \\
BAVS \cite{liu2024bavs} & Swin-B & $224^2$ & & 33.6 & 37.5 & & 82.7 & 89.8 & & 59.6 & 65.9 \\
QDFormer \cite{li2024qdformer} & Swin-T & $224^2$ & & - & - & & 79.5 & 88.2 & & 61.9 & 66.1 \\
DDESeg$^\dagger$$^\ddagger$ \cite{liu2025dynamic} & Trans. & $224^4$ & & 43.3 & 48.7 & & \textbf{89.4} & \textbf{93.2} & & \textbf{67.8} & \textbf{75.1} \\
\textbf{\modelName} & ViT-B & $224^2$ & & \underline{45.8} & \underline{50.7} & & 83.5 & 92.0 & & 61.4 & 70.2 \\
\textbf{\modelName} & ViT-L & $224^2$ & & \textbf{52.2} & \textbf{57.2} & & \underline{85.5} & \underline{92.9} & & \underline{65.0} & \underline{71.1} \\

\midrule

AAVS \cite{ma2024stepping} & Swin-B & $384^2$ & & 48.5 & 53.2 & & 83.2 & 91.3 & & \underline{67.3} & \textbf{77.6} \\
\textbf{\modelName} & ViT-B & $384^2$ & & \underline{49.8} & \underline{54.1} & & \underline{86.2} & \underline{93.3} & & 62.9 & 73.6 \\
\textbf{\modelName} & ViT-L & $384^2$ & & \textbf{56.5} & \textbf{61.0} & & \textbf{87.7} & \textbf{94.1} & & \textbf{70.2} & \underline{76.5} \\

\bottomrule
\end{tabular}
\label{tab:combined_s4_ms3_avss}
\end{table}

\PAR{AVSS Quantitative Comparisons.} 
\Cref{tab:eff_study} shows quantitative comparisons in \ac{AVSS} \cite{zhou2022audio, zhou2024avss} benchmark combined with efficiency statistics.
AVS-Mamba \cite{gong2025avs} or BAVS \cite{liu2024bavs} are excluded from the efficiency evaluation as their source code is not publicly available.
Their \ac{AVSS} segmentation scores are reported in \cref{tab:combined_s4_ms3_avss}, where \ac{\modelName} with ViT-L achieves superior performance. 
For the DDESeg \cite{liu2025dynamic} model, weights are not publicly available, so we train and evaluate the model following the \ac{AVSS} \cite{zhou2022audio, zhou2024avss} standard. 
For COMBO \cite{yang2024cooperation}, we use precomputed maskige representations while evaluating the efficiency.

Using \ac{PVTv2} \cite{wang2022pvt} together with a lightweight audio-visual fusion module, our model outperforms methods with considerably more complex architectures. 
As noted in \cref{ssec:setup}, direct parameter comparisons are complicated by inconsistent audio backbones. 
The smaller parameter count of DDESeg \cite{liu2025dynamic} stems from its smaller audio backbone rather than its visual or fusion design. 
Increasing the input frame size further improves performance while \ac{\modelName} maintains high computational efficiency.

However, \ac{PVTv2} is not a plain \ac{ViT}.
It introduces a hierarchical pyramid structure with overlapping patch embeddings and spatial-reduction attention, which enables multi-scale feature extraction rather than a single-scale global representation. 
Since our approach is not dependent on a specialized architecture, we can utilize a \ac{VFM} and optimization techniques designed for plain attention.

By utilizing a pretrained \ac{ViT} \cite{simeoni2025dinov3}, we achieve \ac{SotA} performance with superior model throughput.
The benefit of large-scale pretraining of \ac{VFM}s is clear in the \ac{AVSS} task, where the class variation is high.
Despite utilizing more parameters with \ac{ViT} \cite{dosovitskiy2020vit} based approaches, \ac{\modelName} achieves superior inference throughput. 
This highlights that the efficiency gains of \ac{\modelName} are architectural in nature, originating from the plain \ac{ViT} design and its compatibility with optimized attention implementations such as FlashAttention \cite{dao2023flashattention2}. 
Although our largest configuration incurs a higher GFLOPs count than the compared baselines, its throughput remains competitive, thanks to modern attention acceleration techniques. 
In applications such as video understanding, robotics, immersive media, and assistive technologies, model throughput is crucial: higher FPS enables real-time operation and smoother integration into larger processing pipelines.
A more comprehensive comparison with existing methods on the AVSS task without efficiency metrics is in \cref{tab:combined_s4_ms3_avss}.

\PAR{S4 and MS3 Quantitative Comparisons.}
With the ViT-L backbone, \ac{\modelName} achieves the best \ac{MS3} Jaccard index and the best \ac{S4} F-score among all compared
specialized methods \cite{zhou2022audio, gao2024avsegformer, yang2024cooperation, ma2024stepping, li2024qdformer, li2024selm, huang2023discovering, chen2024unraveling, li2023catr, liu2024bavs, liu2023audio, gong2025avs, mao2023multimodal, liu2025dynamic}, despite using no task-specific components, as shown in \cref{tab:combined_s4_ms3_avss}.
The results on ViT-B are also highly comparable. 
On the remaining metrics, specialized models retain an advantage in the low-data \ac{S4} and \ac{MS3} regimes. 
We attribute this to the small dataset sizes: the built-in inductive biases of specialized modules, such as multiscale feature extraction, audio source separation, and localized processing, help capture fine-grained spatial detail when training data is very limited. 
Such specialization may also increase the risk of overfitting and, as shown in \cref{tab:eff_study}, reduce overall efficiency.

\subsection{Ablations}

\begin{table}[th]
    \centering
    \caption{
    \textbf{Effect of pretraining on MS3.} 
    We pretrain \modelName~with PVTv2 \cite{wang2022pvt} on S4 data and fine-tune it using MS3.
    }
    \vspace{-5pt}
    \begin{tabular}{ll l cccc}
        \toprule
        \multicolumn{1}{c}{} & \multicolumn{1}{c}{} & & \multicolumn{1}{c}{} & \multicolumn{1}{c}{} & \multicolumn{2}{c}{\textit{MS3}} \\
        \cmidrule{6-7}
        \multicolumn{1}{l}{\multirow{-2}{*}{Method}} & \multicolumn{1}{l}{\multirow{-2}{*}{Backb.}} & & \multicolumn{1}{l}{\multirow{-2}{*}{Input}} & \multicolumn{1}{l}{\multirow{-2}{*}{Pretrained}} & $\mathcal{J} \uparrow$ & $\mathcal{F} \uparrow$ \\
        \midrule
        \textbf{\modelName} & PVTv2 & & $224^2$ & \tikzxmark & 55.3 & 67.4 \\
        \textbf{\modelName} & PVTv2 & & $224^2$ & \tikzcmark & 60.8 & 68.7 \\ 
        \textbf{\modelName} & PVTv2 & & $384^2$ & \tikzxmark & 59.4 & 70.1 \\
        \textbf{\modelName} & PVTv2 & & $384^2$ & \tikzcmark & 61.9 & 69.8 \\
        \bottomrule
    \end{tabular}
    \label{tab:pretraining}
\end{table}
\PAR{Effect of Pretraining on \ac{MS3}.} 
Since our approach does not rely on architectural inductive biases and \ac{PVTv2} \cite{wang2022pvt} is less extensively pretrained than DINOv3 \cite{simeoni2025dinov3}, we first pretrain the \ac{PVTv2}-based model on the \ac{S4} task. 
This stage enables the model to learn basic audio-visual segmentation concepts before fine-tuning on the limited \ac{MS3} dataset. 
By pretraining the model with a $224 \times 224$ input size, we gain approximately 9\% and 2\% performance increases in the Jaccard index and the F-score, respectively. 
For the $384 \times 384$ input model, pretraining increases the Jaccard index by roughly 4\%, but reduces the F-score by approximately 0.4\%.

\begin{table}[th]
    \centering
    \caption{ \textbf{Ablation on different audio enhancers.} \modelName~performance
    using DINOv3-B \cite{simeoni2025dinov3} $224^{2}$ model on AVSBench-Semantic
    \cite{zhou2024avss} benchmark.}
    \vspace{-5pt}
    \begin{tabular}{l l cc}
        \toprule \multicolumn{1}{c}{}                             &  & \multicolumn{2}{c}{\textit{AVSS}} \\
        \cmidrule{3-4}
        \multicolumn{1}{l}{\multirow{-2}{*}{Audio Enhancer Type}} &  & $\mathcal{J}\uparrow$            & $\mathcal{F}\uparrow$ \\
        \midrule Linear                                           &  & 44.3                             & 49.3                  \\
        Precomputed Clusters                                      &  & 43.9                             & 48.8                  \\
        Learned Clusters                                          &  & \textbf{45.8}                    & \textbf{50.7}         \\
        \bottomrule
    \end{tabular}
    \label{tab:abl_audio_enhancer}
\end{table}

\PAR{Audio Enhancer.} 
To assess the impact of audio feature enhancement, we evaluate three configurations: our learned clustering, the pre-calculated audio feature bank enhancement from DDESeg \cite{liu2025dynamic}, and a single linear projection layer as a baseline.
The learned clustering-based audio enhancement significantly outperforms the alternatives.
The pre-calculated audio feature bank from DDESeg \cite{liu2025dynamic} does not improve over the single linear projection baseline, and in some cases reduces performance, suggesting its static nature cannot capture the context-dependent audio-visual interactions in segmentation.

\begin{table}[ht]
    \centering
    \caption{ \textbf{Ablation on different Audio-Visual Fusion Modules.} \modelName~performance
    using DINOv3-B \cite{simeoni2025dinov3} $224^{2}$ model on AVSBench-Semantic
    \cite{zhou2024avss} benchmarks. }
    \vspace{-5pt}
    \begin{tabular}{l l cc}
        \toprule \multicolumn{1}{c}{}                            &  & \multicolumn{2}{c}{\textit{AVSS}} \\
        \cmidrule{3-4}
        \multicolumn{1}{l}{\multirow{-2}{*}{Fusion Module Type}} &  & $\mathcal{J}\uparrow$            & $\mathcal{F}\uparrow$ \\
        \midrule Bi-way Attention                                &  & 43.5                             & 48.7                  \\
        $\mathrm{AVFusion}$ w/o Clustering                       &  & 44.4                             & 49.5                  \\
        $\mathrm{AVFusion}$                                      &  & \textbf{45.8}                    & \textbf{50.7}         \\
        \bottomrule
    \end{tabular}
    \label{tab:abl_fusion}
\end{table}

\PAR{Audio-Visual Fusion.} 
We further ablate the design of the audio-visual fusion module. 
\Cref{tab:abl_fusion} compares three fusion configurations: a simple cross-attention baseline, the bi-directional attention module from CAVP~\cite{chen2024unraveling}, and the clustering-based bi-directional fusion used in \ac{\modelName}, which is inspired by DDESeg \cite{liu2025dynamic}.
The clustering-based fusion module significantly outperforms the alternatives, effectively capturing the complex relationships between the modalities.

\section{Conclusion}
\label{sec:conclusion}
We presented \ac{\modelName}, a simplified encoder-only architecture that removes task-specific complexity from \ac{AVSS} models in favor of a plain \ac{ViT} backbone.
By combining learned audio enhancement with a lightweight fusion module, \ac{\modelName} achieves \ac{SotA} accuracy at 3$\times$ the speed of prior methods.
\ac{\modelName} shows \ac{AVS} can exploit modern \ac{VFM}s and offers a simpler, scalable foundation for future research.


\bibliographystyle{IEEEbib}
\bibliography{bibliography}

\end{document}